\pdfoutput=1
\documentclass[letterpaper]{article} 
\usepackage{aaai25}  
\usepackage{times}  
\usepackage{helvet}  
\usepackage{courier}  
\usepackage[hyphens]{url}  
\usepackage{graphicx} 
\usepackage{natbib}  
\usepackage{caption} 
\usepackage{algorithm}
\usepackage{algorithmic}

\usepackage{amsmath}
\usepackage{amssymb}

\usepackage{multirow}
\usepackage{lscape}
\usepackage{array}
\usepackage{amsfonts} 
\usepackage{mathrsfs}
\usepackage{booktabs}
\usepackage{makecell}

\usepackage{colortbl}

\usepackage{newfloat}
\usepackage{listings}
\DeclareCaptionStyle{ruled}{labelfont=normalfont,labelsep=colon,strut=off} 
\floatstyle{ruled}
\newfloat{listing}{tb}{lst}{}
\floatname{listing}{Listing}
\title{FAMNet: Frequency-aware Matching Network for Cross-domain Few-shot Medical Image Segmentation}
\author{
        Yuntian Bo, 
        Yazhou Zhu, 
        Lunbo Li, 
        Haofeng Zhang\thanks{Corresponding author.}
}
\affiliations{
    School of Computer Science and Engineering, Nanjing University of Science and Technology, China\\
    \{yuntian.bo, zyz\_nj, lunboli, zhanghf\}@njust.edu.cn
}

\usepackage{bibentry}

\begin{document}

\maketitle

\begin{abstract}

Existing few-shot medical image segmentation (FSMIS) models fail to address a practical issue in medical imaging: the domain shift caused by different imaging techniques, which limits the applicability to current FSMIS tasks. To overcome this limitation, we focus on the cross-domain few-shot medical image segmentation (CD-FSMIS) task, aiming to develop a generalized model capable of adapting to a broader range of medical image segmentation scenarios with limited labeled data from the novel target domain.
Inspired by the characteristics of frequency domain similarity across different domains, we propose a Frequency-aware Matching Network (FAMNet), which includes two key
components: a Frequency-aware Matching (FAM) module and a Multi-Spectral Fusion (MSF) module. The FAM module tackles two problems during the meta-learning phase: 1) intra-domain variance caused by the inherent support-query bias, due to the different appearances of organs and lesions, and 2) inter-domain variance caused by different medical imaging techniques. Additionally, we design an MSF module to integrate the different frequency features decoupled by the FAM module, and further mitigate the impact of inter-domain variance on the model's segmentation performance.
Combining these two modules, our FAMNet surpasses existing FSMIS models and Cross-domain Few-shot Semantic Segmentation models on three cross-domain datasets, achieving state-of-the-art performance in the CD-FSMIS task. Code is available at \url{https://github.com/primebo1/FAMNet}.
\end{abstract}

\section{Introduction}
\label{sec:intro}

To bridge the gap between limited labeled samples and the need for precise segmentation, few-shot medical image segmentation (FSMIS) \cite{SENET,ssl-alp,ADNET,zhu2023few,GCN-DE,MRrNet,qnet,CATNET,CRAPNet,GMRD} has emerged. By training on base categories, FSMIS models can leverage only a few annotated samples to segment new categories in medical images directly. Nevertheless, due to the limited generalization capability, they often exhibit diminished performance when tested on the data with domain shifts, which restricts their applicability to only a single domain.

Recently, some researchers have started investigating cross-domain few-shot semantic segmentation (CD-FSS) \cite{PATNET,PMNET,ABCD,DRADAPTER,IFA,APseg}, which has demonstrated impressive segmentation capabilities on datasets like Deepglobe \cite{DeepGlobe} and FSS-1000 \cite{FSS1000}. Although this operation paves the way for cross-domain applications in few-shot scenarios, these models cannot be directly applied to the medical field due to the unique characteristics of medical images, \textit{e.g.}, grayscale, intensity variations, and foreground-background imbalance. Meanwhile, existing domain generalization methods in medical imaging \cite{ouyang2021causality,zhou2022dn,10.1007/978-3-031-16449-1_64,Su_Yao_Yang_Huang_Wang_Sun_2023} mainly focus on domain randomization, neglecting the model itself and the few-shot setting.

\begin{figure}[t]
    \centering
    \includegraphics[width=\linewidth]{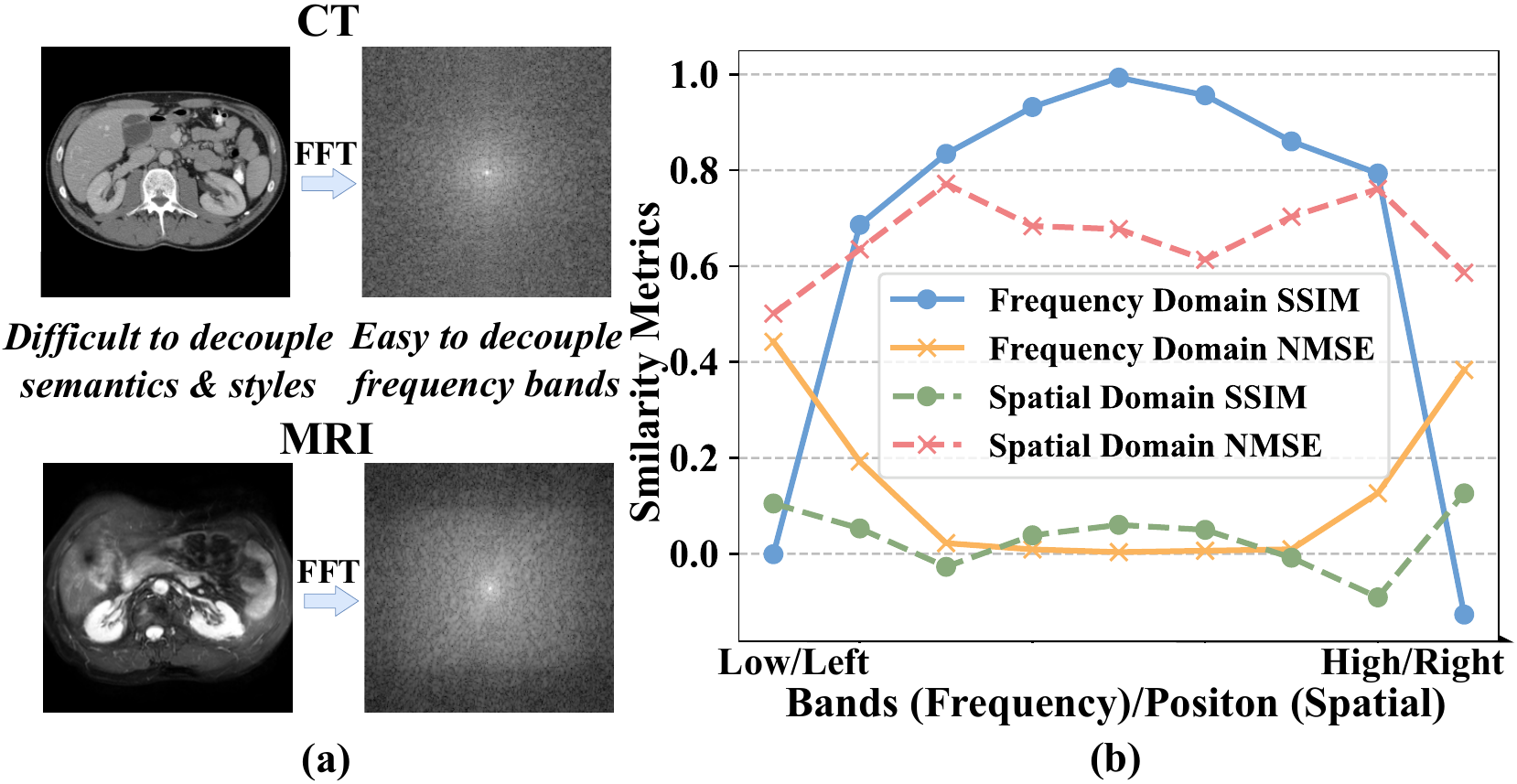}
    \caption{Motivation of the proposed method. (a) CT and MRI scans in the spatial and frequency domains. Frequency spectra are processed using a Hamming window \cite{HAMMING} and are center-shifted. (b) Quantitative metrics for the similarity of CT and MRI in the spatial and frequency domains using structural similarity index measure (SSIM) \cite{SSIM} and normalized mean square error (NMSE). Metrics are calculated using registered images.}
    \label{fig:motivation}
\end{figure}

Two major challenges hinder the development of cross-domain few-shot medical image segmentation (CD-FSMIS), which we try to address in this paper:
1)	Intra-domain variations: Medical images exhibit significant variability between individual organs, \textit{e.g.}, size, fat content, and pathology, making it difficult to find similar support-query pairs, leading to support-query bias and reduced prototype representation in prototypical networks.
2)	Inter-domain variations: 

Even within the same organ or region, the spatial domain similarity demonstrates low correlation across different domains, as illustrated in Figure \ref{fig:motivation}(b). However, subtle distinctions are evident in the frequency domain, where inter-domain variations are primarily in high and low-frequency bands, while mid-frequency bands are relatively similar.

We therefore propose a novel method termed Frequency-aware Matching Network (FAMNet) for CD-FSMIS in this paper. Specifically, the core of our FAMNet, the Frequency-aware Matching (FAM) module, performs support-query matching in specific frequency bands, eliminating support-query bias by fusing foreground features and highlighting synergistic parts. Simultaneously, FAM incorporates frequency domain information within the feature space, reducing reliance on frequency bands with significant domain differences. This allows the model to focus more on resilient, domain-agnostic frequency bands, effectively addressing both intra-domain and inter-domain variations.
Building upon the FAM module, we subsequently developed a Multi-spectral Fusion (MSF) module. While fusing the frequency-decoupled features from FAM, the MSF module extracts the critical information that remains in the domain-specific frequency bands after decoupling in the spatial domain. With FAM and MSF, our method not only demonstrates strong generalization capabilities but also effectively leverages domain-invariant interactive information from the sample space, showcasing excellent segmentation performance. In summary, our contributions are as follows:

\begin{itemize}
    \item We extend few-shot medical image segmentation to a new task, termed cross-domain few-shot medical image segmentation, aimed at training a generalizable model to segment a novel class in unseen target domains with only a few annotated examples.
    \item We propose a novel FAM module that concurrently mitigates the adverse impacts of intra-domain and inter-domain variances on model performance. Moreover, an MSF module is introduced for multi-spectral feature fusion, further suppressing domain-variant information to enhance the model's generalizability.
    \item On three cross-domain datasets, our proposed method archives the state-of-the-art performance.   
    The effectiveness and superiority of our method are further verified through various ablation studies and visualization.
\end{itemize}

\section{Related Works}

\subsection{Few-shot Medical Image Segmentation}
The FSMIS task has been proposed to address data scarcity typically found in medical scenarios, which aims to train models capable of segmenting novel organs or lesions with only a few annotated samples. Current FSMIS models can be categorized into two approaches: interactive networks \cite{SENET,GCN-DE,MRrNet, CRAPNet} and prototypical networks \cite{ssl-alp, ADNET, qnet, zhu2023few, CATNET, GMRD}. In the former category, SENet \cite{SENET} pioneered the use of interactive networks in FSMIS tasks, followed by MRrNet \cite{MRrNet}, GCN-DE \cite{GCN-DE}, and CRAPNet \cite{CRAPNet}. The core idea behind these models is to enhance support-query interaction through attention mechanisms. For the latter category, SSL-ALPNet \cite{ssl-alp} introduced a self-supervised framework that generates adaptive local prototypes and supervised by superpixel-based pseudo-labels during training. ADNet \cite{ADNET} proposed a learnable threshold for segmentation and relied on a single foreground prototype to compute anomaly scores for all query pixels, rather than learning prototypes for each class. CATNet \cite{CATNET} utilized a cross-masked attention Transformer to enhance support-query interaction and improve feature representation. GMRD \cite{GMRD} captured the complexity of prototype class distributions by generating multiple representative descriptors. Unfortunately, all existing FSMIS methods are limited to single-domain applications, neglecting the domain shifts encountered in medical imaging.

\subsection{Cross-domain Few-shot Semantic Segmentation}
Expanding on few-shot semantic segmentation (FSS), recent studies \cite{ABCD,APseg,DRADAPTER,IFA,PMNET,PATNET} focus on CD-FSS, considering a more practical setting where both label space and data distribution are disjoint between the training and testing datasets. PATNet \cite{PATNET} employs a Pyramid-Anchor-Transformation module (PATM) to map domain-specific features into domain-agnostic ones. PMNet \cite{PMNET} proposes a lightweight matching network to densely exploit pixel-to-pixel and pixel-to-patch correlations between support-query pairs. DRAdapter \cite{DRADAPTER} utilizes local-global style perturbation to train an adapter that rectifies diverse target domain styles to the source domain, maximizing the utilization of the well-optimized source domain segmentation model.
Nevertheless, existing CD-FSS models often suffer from substantial performance degradation when applied to medical images due to significant differences from natural images, such as color, intensity, and foreground-background imbalance.

\section{Methodology}

\begin{figure*}[t]
\centering
    \includegraphics[width=0.97\linewidth]{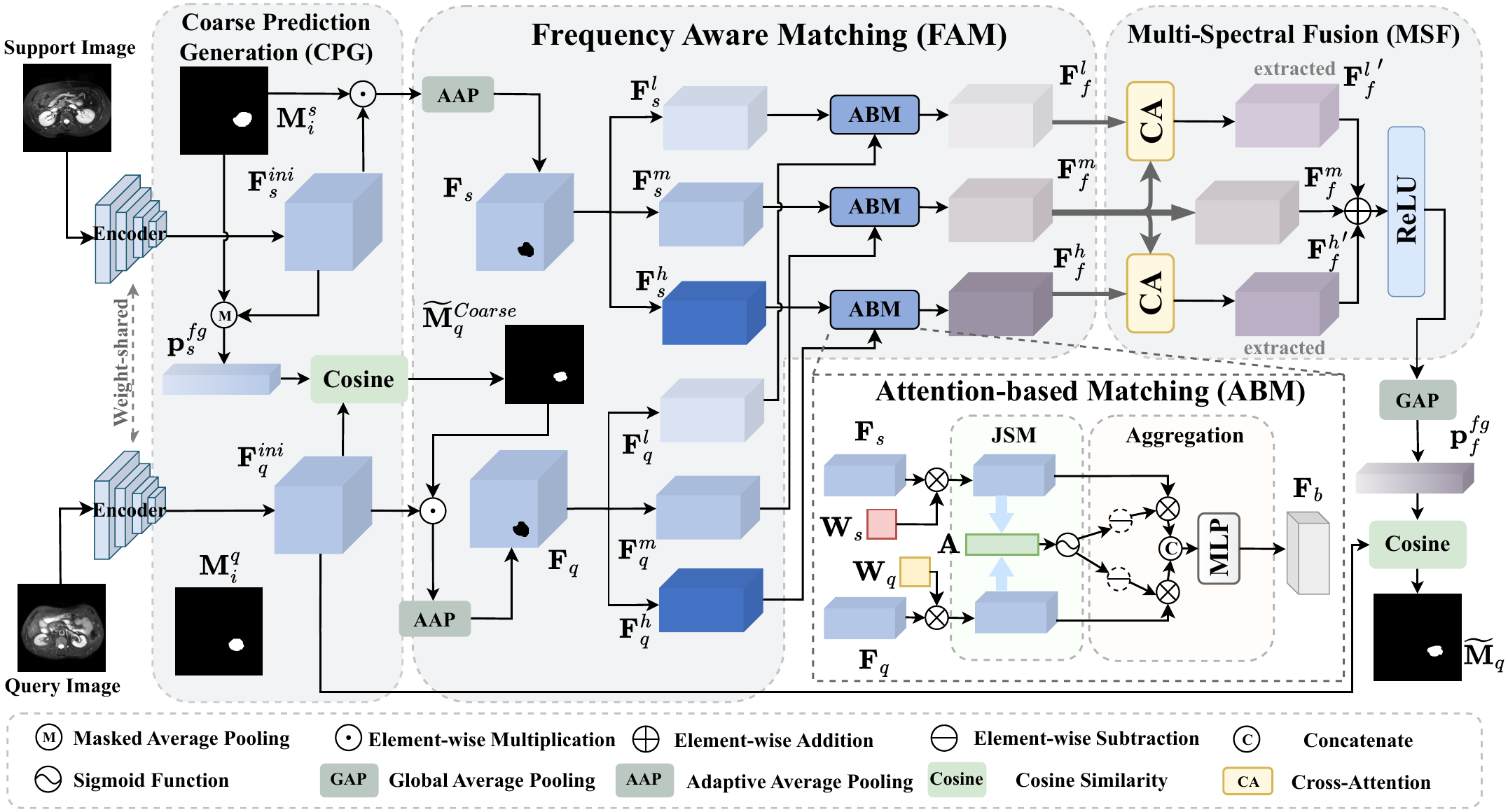}
    \caption{The overall architecture of our method, consists of three main technical components: the Coarse Prediction Generation (CPG) module, the Frequency-aware Matching (FAM) module, and the Multi-Spectral Fusion (MSF) module. Note that in ABM, JSM denotes joint space matching. In the case of DAFBs, the attention matrix is directly utilized for attention weighting. Conversely, for DSFBs, an element-wise subtraction is applied prior to the weighting process. }
    \label{fig:overall architecture}
\end{figure*}

\subsection{Problem Setting}

The CD-FSMIS task aims to construct a generalizable model $\Theta$ to segment novel organs or lesions in an unseen domain with few annotated medical images. To elaborate, the model $\Theta$ is optimized using a single source domain dataset $\mathcal{D}^{s}$ encompassing the base categories $\mathcal{C}_{base}$. Subsequently, the model's performance is assessed on a target domain dataset $\mathcal{D}^{t}$, which comprises novel target categories $\mathcal{C}_{target}$ with only a few labeled images. It is crucial to highlight that the sets of categories $\mathcal{C}_{base}$ and $\mathcal{C}_{target}$ are disjoint, \textit{i.e.} $\mathcal{C}_{base} \cap \mathcal{C}_{target} = \emptyset$, and a domain shift exists between the source domain $\mathcal{D}^{s}$ and the target domain $\mathcal{D}^{t}$. During the training phase, the model has no access to the target domain.
 
Our approach adheres to the episode-based meta-learning paradigm. For each meta-learning task, we randomly divide the data into multiple episodes. Each episode $(\mathcal{S}, \mathcal{Q})$ comprises: 1) a support set  $\mathcal{S} = \{x_i^s, \mathbf{M}_i^s\}_{i=1}^{K}$ containing $K$ support samples, and 2) a query set $\mathcal{Q} = \{x_i^q, \mathbf{M}_i^q\}_{i=1}^{N_q}$ containing $N_q$ query samples, where $x_i$ denotes the $i$-th image, and $\mathbf{M}_i$ denotes the corresponding segmentation ground truth. During inference, the model's segmentation performance is evaluated by providing a support set and a query set from the novel target domain.

\subsection{Overall Architecture}
The proposed network is depicted in Figure \ref{fig:overall architecture}, which can be briefly divided into three main parts: 1) a Coarse Prediction Generation (CPG) module for generating a coarse prediction of the query mask, 2) a Frequency-aware Matching (FAM) module for performing frequency-aware matching between support and query foreground features, and 3) a Multi-Spectral Fusion (MSF) module for fusing features based on their respective frequency bands.

First, the support and query images are fed into a weight-sharing feature encoder to extract their corresponding feature maps. Next, a coarse prediction of the query foreground mask is obtained using the CPG module. Then, the support foreground feature computed in CPG, the generated coarse query mask, and the extracted query feature are input into the proposed FAM module for frequency-aware matching. In this module, features are divided into three frequency bands, each independently fuses the support and query features through distinct weighting mechanisms guided by matching results, yielding fused features for each band. To reintegrate these multi-spectral features and further suppress the influence of domain-specific frequency bands (DSFBs), the divided features are fused through the MSF module. A global average pooling operation is then performed to obtain the foreground prototype required by the prototypical network. Finally, we compute the cosine similarity between the foreground prototype and the query feature to produce the final prediction of the query mask.

\subsection{Feature Extraction}
We use a ResNet-50 \cite{7780459} feature encoder $\mathcal{E}_{\theta}(\cdot)$  pre-trained on MS-COCO \cite{10.1007/978-3-319-10602-1_48} as a weight-shared backbone to extract the support and query feature maps, where $\theta$ denotes the backbone parameters. 

The support and query feature maps are denoted as $\mathbf{F}_s^{ini} = \mathcal{E}_{\theta}(x_i^s)$ and $\mathbf{F}_q^{ini} = \mathcal{E}_{\theta}(x_i^q)$, and $\mathbf{F}_s^{ini}$, $\mathbf{F}_q^{ini} \in \mathbb{R}^{C \times H \times W}$, where $C$ denotes the channel depth of the feature, and $H$ and $W$ denote the height and width of the feature respectively.

\subsection{Coarse Prediction Generation (CPG) Module}

A prototypical network-based method is performed to obtain a coarse segmentation mask of the query image. Given a support image $x_i^s$ and its corresponding foreground binary mask $\mathbf{M}^s$, the support foreground prototype $\mathbf{p}_s^{fg}$ can be generated by using Masked Average Pooling (MAP). Mathematically, this process can be denoted as:
\begin{equation}
\mathbf{p}_s^{fg} =  \frac{\sum_{u, v} \mathbf{F}_s^{ini}{(u, v)}\mathbf{M}^s{(u, v)}}{\sum_{u, v}  \mathbf{M}^s{(u, v)}},
\end{equation}
where $(u, v)$ is the index of pixels on the feature map.

Following this, we directly use the computed $\mathbf{p}_s^{fg}$ and the extracted query feature map $\mathbf{F}_q^{ini}$ to predict a coarse query foreground mask $\widetilde{\mathbf{M}}_{q}^{Coarse}$: 

\begin{equation}
\label{eq:mask}
\widetilde{\mathbf{M}}_{q}^{Coarse} = 1.0 - \sigma\left(d(\mathbf{F}_q^{ini}, \mathbf{p}_s^{fg}) - \tau \right),
\end{equation}
where $d(a, b) = -\alpha cos(a, b)$ is the negative cosine similarity with a fixed scaling factor $\alpha$ = 20 \cite{PANET}, $\sigma(\cdot)$ denotes the Sigmoid activation and $\tau$ is a learnable threshold introduced by \cite{ADNET}.

\subsection{Frequency-Aware Matching (FAM) Module}

\textbf{Multi-Spectral Decoupling of Foreground Features.}
 To mitigate the discrepancy between support and query foregrounds, we first extract the foreground features from $\mathbf{f}_s$ and $\mathbf{f}_q$ using the support mask and coarse query mask. Since the number of foreground pixels in the support and query images often differs, we apply Adaptive Average Pooling (AAP) \cite{AAP} to standardize the number of foreground pixels to a fixed value $N$. This process can be formulated as: 
\begin{equation}
\left\{
\begin{array}{ll}
    \mathbf{F}_s= \text{AAP}(\mathbf{F}_s^{ini} \odot \mathbf{M}^s, N) \\
    \mathbf{F}_q = \text{AAP}(\mathbf{F}_q^{ini} \odot \mathcal{R}(\widetilde{\mathbf{M}}_{q}^{Coarse}), N)
\end{array}
\right.,
\end{equation}
where $\odot$ denotes the Hadamard product, and $\mathbf{F}_s$, $\mathbf{F}_q \in \mathbb{R}^{C \times N}$ denote the extracted foreground features, and $\text{AAP}(a, n)$ is the AAP operation that adjusts the input feature map $a$ to a fixed output size $n$ along the last dimension, and $\mathcal{R}$ denotes the mathematical function that rounds decimals to 0 or 1.

For each foreground feature, we utilize the two-dimensional Fast Fourier Transform (FFT) to convert the signal from the spatial domain to the frequency domain while preserving spatial information. We employ a reshape function $\rho$ to transform the feature into a square, $\rho: \mathbb{R}^{C \times N} \to \mathbb{R}^{C \times \sqrt{N} \times \sqrt{N}}$, with $\rho^{-1}$ serving as its inverse. For the support foreground feature, this process is formalized as:
\begin{equation}
\phi_s = \text{SC}(\mathcal{F}(\rho(\mathbf{F}_s))),
\end{equation}
where $\mathcal{F}$ denotes FFT, $\phi_s \in \mathbb{C}^{C \times \sqrt{N} \times \sqrt{N}}$ denotes the frequency domain feature representation, $\text{SC}$ denotes the function to adjusts the frequency signal to center the zero-frequency component.

Subsequently, we apply a band-pass filter to decompose the frequency-domain signal into three bands, namely high, medium, and low frequencies. Finally, we revert the frequency signals back to the spatial domain using the Inverse Fast Fourier Transform (IFFT): 
\begin{equation}
\mathbf{F}_s(i) = \rho^{-1}(\mathcal{F}^{-1}(\mathcal{B}_p(\phi_s, I(i)))) = 
\begin{cases}
\mathbf{F}_s^l, &  i = 1 \\
\mathbf{F}_s^m, &  i = 2 \\
\mathbf{F}_s^h, &  i = 3
\end{cases},
\end{equation}
where $\mathbf{F}_s(i)$ denotes the support foreground feature in a specific frequency band,  $\mathcal{F}^{-1}$ denotes  IFFT, $I(i)$ is a binary mask vector where values are set to 1 for preserved components and 0 for discarded components, and $\mathcal{B}_p(\phi, I)$ is the band-pass filter that can be defined by:
\begin{equation}
\phi'(u, v) = \begin{cases}
0, & \text{if } I(u, v) = 0 \\
\phi(u, v), & \text{otherwise}
\end{cases}.
\end{equation}

We perform the same operation on the query foreground feature to obtain $\mathbf{F}_q^l$, $\mathbf{F}_q^m$, $\mathbf{F}_q^h$. \\ 
\textbf{Multi-Spectrum Attention-based Matching.}
As shown in Figure \ref{fig:motivation}, high-frequency and low-frequency signals vary significantly across different domains. During typical training, models often rely on these DSFBs to better adapt to the tasks in the current scenario. However, this reliance leads to over-fitting to certain prominent features. For example, a model may achieve precise segmentation on CT images by focusing on the contained information of DSFBs, such as the high-frequency edge information. Yet, this approach often suffers from degradation when transferred to an unseen domain where edge information is less distinct, such as MRI. Our proposed method aims to enhance the model's generalization capabilities by matching support and query features while simultaneously reducing the model's reliance on DSFBs.

Specifically, this module consists of three processes, as illustrated in the top-right part of Figure \ref{fig:overall architecture}. First, given the support and query foreground feature pair  ($\mathbf{f}_s$, $\mathbf{f}_q$) belonging to the same frequency band $\mathcal{B}$, we apply linear transformations using two learnable matrices $\mathbf{W}_s$, $\mathbf{W}_q \in \mathbb{R}^{N \times N}$ to map the features into a joint space. This further reduces the intra-domain differences between the support and query, enhancing the stability of matching. This operation can be formalized as:
\begin{equation}
\left\{
\begin{aligned}
\mathbf{F}_s' &= \mathbf{F}_s\mathbf{W}_s \\
\mathbf{F}_q' &= \mathbf{F}_q\mathbf{W}_q
\end{aligned}
\right.,
\end{equation}
where $\mathbf{F}_s'$, $\mathbf{F}_q' \in \mathbb{R}^{C \times N}$ denote the transformed support and query foreground features, respectively.

Secondly, based on the transformed features, we compute the attention-based similarity score matrix between the features using cosine similarity:
\begin{equation}
\mathbf{A}(\mathbf{F}_s', \mathbf{F}_q') = \sigma(\frac{\mathbf{F}_s' \cdot \mathbf{F}_q'}{\|\mathbf{F}_s'\| \|\mathbf{F}_q'\|}),
\end{equation}
where $\mathbf{A} \in \mathbb{R}^{1 \times N}$ denotes the computed attention matrix, and $\sigma$ denotes the sigmoid activation function.
During training, our module updates the attention scores between features via the learnable transformation matrices $\mathbf{W}_s$ and $\mathbf{W}_q$, enabling the model to better learn domain-agnostic similarities between features.

Thirdly, based on the notion that DSFBs and domain-agnostic frequency bands (DAFBs) should be treated differently, we apply distinct attention weightings to feature pairs in three frequency bands. To match the size of $\mathbf{F}_s'$ and $\mathbf{F}_q'$, we first obtain $\mathbf{A}' \in \mathbb{R}^{C \times N}$ by repeating $\mathbf{A}$ along the channel dimension. For feature pairs in DAFBs, We directly multiply the attention matrix with the features element-wise to highlight the similar components while suppressing the dissimilar ones: 
\begin{equation}
\left\{
\begin{aligned}
\mathbf{F}_{s,0}'' &= \mathbf{A}' \odot \mathbf{F}_s' \\
\mathbf{F}_{q,0}'' &= \mathbf{A}' \odot \mathbf{F}_q'
\end{aligned}
\right.,
\end{equation}
where $\mathbf{F}_{s,0}''$ and $\mathbf{F}_{q,0}'' \in \mathbb{R}^{C \times N}$ are the weighted features belonging to the DAFBs.

While for feature pairs in DSFBs, an inverse attention weighting is performed to suppress the similar components between features, thereby reducing the model's reliance on these components:
\begin{equation}
\left\{
\begin{aligned}
\mathbf{F}_{s,1}'' &= (1 - \mathbf{A}') \odot \mathbf{F}_s' \\
\mathbf{F}_{q,1}'' &= (1 - \mathbf{A}') \odot \mathbf{F}_q' 
\end{aligned}
\right. ,
\end{equation}
where $\mathbf{F}_{s,1}''$ and $\mathbf{F}_{q,1}'' \in \mathbb{R}^{C \times N}$ are the weighted features belonging to the DSFBs.

Thus, the FAM module completes the enhancement or suppression of full-spectrum similarity between features through $\mathbf{A}$, which is derived from full-spectrum attention weights. We pass the concatenated feature pair through an MLP afterward to fuse the features belonging to the support and query, obtaining a new fused feature that acts as the final feature representation of $\mathcal{B}$: 
\begin{equation}
\mathbf{F}_b = \text{MLP}\left( \text{Cat}\left(\mathbf{F}_{s,fb}'', \mathbf{F}_{q,fb}''\right), \varphi \right), \quad fb \in \{0, 1\}
\end{equation}
where $\mathbf{F}_b \in \mathbb{R}^{C \times N}$ denotes the fused feature in a specific frequency band,  $\text{MLP}(\cdot, \varphi)$ is a function of a MLP with parameters $\varphi$. Notably, the MLP consists of two fully connected layers with a ReLU activation function in between, which can better learn the fusion patterns and reduce the parameters of the MLP, Cat($a, b$) denotes the function that concatenates $a$ and $b$ along the last dimension, and $fb$ serves as an indicator specifying DAFBs or DSFBs.

\begin{table*}[!t]
\centering
\begin{tabular}{c|c|ccccc|ccccc}
\toprule
\rule{0pt}{2.5ex}
\multirow{2}{*}{Method} & \multirow{2}{*}{Ref.} & \multicolumn{5}{c|}{\textbf{Abdominal CT $\to$ MRI}} & \multicolumn{5}{c}{\textbf{Abdominal MRI $\to$ CT}} \\ 
\cline{3-12}
\rule{0pt}{2.5ex}

                        &                       & \multicolumn{1}{c|}{Liver} & \multicolumn{1}{c|}{LK} & \multicolumn{1}{c|}{RK} & \multicolumn{1}{c|}{Spleen} & Mean  & \multicolumn{1}{c|}{Liver} & \multicolumn{1}{c|}{LK} & \multicolumn{1}{c|}{RK} & \multicolumn{1}{c|}{Spleen} & Mean  \\ \hline
\rule{0pt}{2.5ex}
 PANet                  & ICCV'19               &                             39.24&                                26.47&                           37.35&         26.79&                      32.46&                        40.29&                    30.61&               26.66&   30.21& 31.94\\
 SSL-ALP                & TMI'22                &                             \underline{70.74}&                                \underline{55.49}&                           \underline{67.43}&         \textbf{58.39}&                      \underline{63.01}&                        71.38&                    34.48&               32.32&   51.67&47.46\\
 ADNet                  & MIA'22                &                             50.33&                                39.36&                           37.88&         39.37&                      41.73&                        64.25&                    37.39&               25.62&   42.94& 42.55\\
 QNet                   & IntelliSys'23         &                             58.82&                                42.69&                           51.67&         44.58&                      49.44&                        70.98&                    38.64&               30.17&   43.28& 45.77\\
 CATNet                 & MICCAI'23             &                             44.58&                                43.67&                           50.27&         46.34&                      46.21&                        54.52&                    41.73&               40.24&   45.84& 45.60\\
 RPT                    & MICCAI'23             &                             49.22&                                42.45&                           47.14&         48.84&                      46.91&                        65.87&                    40.07&               35.97&   51.22& 48.28\\
 PATNet                 & ECCV'22               &                             57.01&                                50.23&                           53.01&         51.63&                      52.97&                        \textbf{75.94}&                    \underline{46.62}&               \underline{42.68}&   \underline{63.94}& \underline{57.29}\\ 
 IFA                    & CVPR'24               &                             48.81&                                45.79&                           51.46&         51.42&                      49.37&                        50.05&                    36.45&               32.69&   43.08& 40.57\\   \hline
 \rule{0pt}{2.5ex}
 \textbf{Ours}        & \textemdash    &\textbf{73.01}              &\textbf{57.28}               & \textbf{74.68}          &\underline{58.21}       &\textbf{65.79}                           &\underline{73.57}&\textbf{57.79}               & \textbf{61.89}          &\textbf{65.78}       &\textbf{64.75}       \rule{0pt}{2.5ex}\\ \bottomrule
\end{tabular}
\caption{Quantitative comparison of different methods Dice score (\%) on the Cross-Modality Dataset. The best value is shown in bold font, and the second best is underlined.}
\label{ABD}
\end{table*}

\begin{table*}[!t]
\centering
\begin{tabular}{c|c|cccc|cccc}
\toprule
\rule{0pt}{2.5ex}
\multirow{2}{*}{Method} & \multirow{2}{*}{Ref.} & \multicolumn{4}{c|}{\textbf{Cardiac LGE $\to$ b-SSFP }} & \multicolumn{4}{c}{\textbf{Cardiac b-SSFP $\to$ LGE}} \\ 
\cline{3-10} 
\rule{0pt}{2.5ex}
                        &                       & \multicolumn{1}{c|}{LV-BP} & \multicolumn{1}{c|}{LV-MYO} & \multicolumn{1}{c|}{RV} & Mean  & \multicolumn{1}{c|}{LV-BP} & \multicolumn{1}{c|}{LV-MYO} & \multicolumn{1}{c|}{RV} & Mean  \\ \hline
\rule{0pt}{2.5ex}
 PANet                  & ICCV'19               & 51.43& 25.75& 25.75& 36.66& 36.24                      & 26.37                       & 23.47                   & 28.69 \\
 SSL-ALP                & TMI'22                & 83.47& 22.73& 66.21& 57.47& 65.81& 25.64& 51.24& 47.56\\
 ADNet                  & MIA'22                & 58.75                      & 36.94                       & 51.37                   & 49.02 & 40.36                      & 37.22                       & 43.66                   & 40.41 \\
 QNet                   & IntelliSys'23         & 50.64                      & 37.88                       & 45.24                   & 44.58 & 31.08                      & 34.03                       & 39.45                   & 34.85 \\
 CATNet                 & MICCAI'23             & 64.63                      & 42.41                       & 56.13                   & 54.39 & 45.77                      & 43.51                       & 46.02                   & 45.10 \\
 RPT                    & MICCAI'23             & 60.84                      & 42.28                       & 57.30                   & 53.47 & 50.39                      & 40.13                       & 50.50                   & 47.00 \\
 PATNet                 & ECCV'22               & \underline{65.35}& \underline{50.63}           & 68.34& 61.44& 66.82& \textbf{53.64}                       & \underline{59.74}& \underline{60.06} \\ 
 IFA                    & CVPR'24               & 64.04& 43.22& \underline{74.58}& \underline{62.28}& \underline{68.07}& 36.07& \textbf{60.42}& 54.85\\ \hline
 \rule{0pt}{2.5ex}
 \textbf{Ours}        &\textemdash             & \textbf{86.64}& \textbf{51.84}& \textbf{76.26}& \textbf{71.58}&\textbf{77.37}&\underline{52.05}&54.75& \textbf{61.39}      \rule{0pt}{2.5ex}\\ \bottomrule
\end{tabular}
\caption{Quantitative comparison of different methods Dice score (\%) on the Cross-Sequence Dataset. The best value is shown in bold font, and the second best is underlined.}
\label{CARDIAC}
\end{table*}

\subsection{Multi-Spectral Fusion (MSF) Module}
\label{MSF module}
Recent research and our experiments in the Supplementary Materials have revealed the subtle relationship between frequency domain signals and image feature information: 1) Different frequency bands contain different information. Low and high frequencies contain color and style information, while the middle-frequency band contains more structural and shape information \cite{Huang_2021_CVPR}. 2) In cross-domain tasks, low and high frequencies exhibit significant differences across different domains. 3) Directly discarding high and low-frequency features is unreasonable, as frequency domain decomposition fails to completely decouple domain-variant information (DVI) and domain-invariant information (DII).

These insights lead us to a question: \textit{Is there a mechanism that can extract the DII remaining in the DSFBs guided by the DII in the DAFBs?} We thus considered the cross-attention mechanism, which is widely used in multi-modal feature fusion: When feature $\mathbf{\Phi}$ is used as the key ($\mathbf{K}$) and value ($\mathbf{V}$), and another feature $\mathbf{\Psi}$ is used as the query ($\mathbf{Q}$), the resulting $\mathbf{V}$ is the representation of feature $\mathbf{\Phi}$ weighted by the similarity between feature $\mathbf{\Psi}$ and feature $\mathbf{\Phi}$.

Propelled by this knowledge, we propose the MSF module, a cross-attention-based feature fusion module. This module retains high and low-frequency information while using mid-frequency information to extract DII from high and low-frequency features and suppress DVI.

To be specific, for each feature triplet ($\mathbf{F}_f^l$, $\mathbf{F}_f^m$, $\mathbf{F}_f^h$), the three fused features do not overlap in the frequency domain, \textit{i.e.}, $\phi(\mathbf{F}_f^l) \cap \phi(\mathbf{F}_f^m) = \varnothing$, $\phi(\mathbf{F}_f^l) \cap \phi(\mathbf{F}_f^h) = \varnothing$, and $\phi(\mathbf{F}_f^m) \cap \phi(\mathbf{F}_f^h) = \varnothing
$, where $\phi(\mathbf{F})$ represents the spectrum of $\mathbf{F}$. However, $\zeta(\mathbf{F}_f^l) \cap \zeta(\mathbf{F}_f^m) \neq \varnothing$ and $\zeta(\mathbf{F}_f^m) \cap \zeta(\mathbf{F}_f^h) \neq \varnothing$, where $\zeta(\mathbf{F})$ denotes the contained information of $\mathbf{F}$. We use $\mathbf{F}_f^l$ or $\mathbf{F}_f^h$, along with $\mathbf{F}_f^m$ to compute the attention between the given features. The output matrix of cross-attention (CA) can be represented as:
\begin{equation}
f_\text{CA}(\mathbf{Q}, \mathbf{K}, \mathbf{V}) = \text{softmax}(\frac{\mathbf{QK}^T}{\sqrt{d}})\mathbf{V} = \mathbf{SV},
\end{equation}
where $d$ denotes a scaling factor, $\mathbf{S} \in \mathbb{R}^{N \times N}$ denotes the attention weight matrix. Consequently, the process of obtaining the refined features for the low and high-frequency components can be formalized by:
\begin{equation}
\left\{
\begin{aligned}
{\mathbf{F}_f^l}' &= f_\text{CA}((\mathbf{F}_f^m)^T \mathbf{W}_Q, (\mathbf{F}_f^l)^T \mathbf{W}_K, (\mathbf{F}_f^l)^T \mathbf{W}_V)^T \\
{\mathbf{F}_f^h}' &= f_\text{CA}((\mathbf{F}_f^m)^T \mathbf{W}_Q, (\mathbf{F}_f^h)^T \mathbf{W}_K, (\mathbf{F}_f^h)^T \mathbf{W}_V)^T
\end{aligned}
\right.,
\end{equation}
where ${\mathbf{F}_f^l}'$, ${\mathbf{F}_f^h}'$ denote the refined fused foreground features, and $\mathbf{W}_Q$, $\mathbf{W}_K$, $\mathbf{W}_V \in \mathbb{R}^{C \times C}$ are the learnable linear transformation matrices. 

Finally, a straightforward addition is performed to integrate the features from the three frequency bands, followed by a \text{ReLU} activation function, as the module output $\mathbf{F}_f$:
\begin{equation}
\mathbf{F}_f = \text{ReLU}({\mathbf{F}_f^l}' + \mathbf{F}_f^m + {\mathbf{F}_f^h}') \in \mathbb{R}^{C \times N}.
\end{equation}

We use the fused foreground features to compute the final query mask. A global average pooling (GAP) operation is performed to obtain the frequency-aware and query-informed foreground prototype:
\begin{equation}
\mathbf{p}_f^{fg}(c) = \frac{1}{N} \sum_{i=1}^{N} \mathbf{F}_f(c, i),
\end{equation}
where $c$ denotes the channel index.

Hence, the final query foreground prediction of our proposed model can be calculated in a similar way in Eq. \ref{eq:mask}:
\begin{equation}
\widetilde{\mathbf{M}}_{q}^{fg} = 1.0 - \sigma\left(d(\mathbf{F}_q^{ini}, \mathbf{p}_f^{fg}) - \tau \right),
\end{equation}
while the background prediction can be obtained by $\widetilde{\mathbf{M}}_{q}^{bg} = 1 - \widetilde{\mathbf{M}}_{q}^{fg}$  accordingly.

\subsection{Objective Function}
We adopt the binary cross-entropy loss $\mathcal{L}_{ce}$
   to evaluate the error between the predicted query mask and its corresponding ground truth. 
   Mathematically, our final prediction loss $\mathcal{L}_{final}$ can be expressed as:
\begin{equation}
\begin{aligned}
\mathcal{L}_{final} &= \mathcal{L}_{ce}(\mathbf{M}^{q}, \widetilde{\mathbf{M}}_{q}^{fg}, \widetilde{\mathbf{M}}_{q}^{bg}) \\
&= -\frac{1}{HW} \sum_{h, w} \mathbf{M}^{q} \log(\widetilde{\mathbf{M}}_{q}^{fg}) + (1 - \mathbf{\mathbf{M}}^{q}) \log(\widetilde{\mathbf{M}}_{q}^{bg}).
\end{aligned}
\end{equation}

To capture more precise and sufficient query foreground features, an accurate coarse prediction of the query foreground is needed. We continue to use the binary cross-entropy loss to quantify the dissimilarity between the coarse prediction and $\mathbf{M}_{q}$:
\begin{equation} 
\mathcal{L}_{coarse} = \mathcal{L}_{ce}(\mathbf{M}_{q}, \widetilde{\mathbf{M}}_{q}^{coarse}, 1 - \widetilde{\mathbf{M}}_{q}^{coarse}).
\end{equation}

Overall, the computation of the total loss $\mathcal{L}_{total}$ for our proposed model can be denoted as $\mathcal{L}_{total} = \mathcal{L}_{final} + \mathcal{L}_{coarse}$.

\section{Experiments}
\subsection{Datasets}
We detail our proposed task into three cross-domain settings and evaluate our method on the following three datasets: 

\textbf{The Cross-Modality dataset} comprises two abdominal datasets. The first is Abdominal MRI obtained from \cite{KAVUR2021101950}, which includes 20 3D T2-SPIR MRI scans. The second is Abdominal CT, which comprises 20 3D abdominal CT scans from \cite{landman2015miccai}. We select four common categories from the two datasets: the left kidney (LK), right kidney (RK), liver, and spleen, for assessment.

\textbf{The Cross-Sequence dataset} is a cardiac dataset from \cite{ZHUANG2022102528}, which includes 45 3D LGE MRI scans and 45 b-SSFP MRI scans, both comprising 3 distinct labels: the blood pool (LV-BP), the left ventricle myocardium (LV-MYO), and the right ventricle myocardium (RV).

\textbf{The Cross-Institution dataset} consists of 321 3D prostate T2-weighted MRI scans collected by the University College London hospitals (UCLH) and 82 3D prostate MRI scans from the National Cancer Institute (NCI), Bethesda, Maryland, USA. The data from UCLH are collected from 4 studies: INDEX \cite{INDEX}, the SmartTarget Biopsy Trial \cite{BIOPSY}, PICTURE \cite{PICTURE}, Promise12 \cite{PICTURE}, and organized by \cite{prototypicalCROSSINSTITUTION}. The data from NCI are provided in \cite{NCI}. All the data are annotated by \cite{prototypicalCROSSINSTITUTION}. We select three common categories, bladder, central gland (CG) and rectum, for assessment.

\subsection{Implementation Details}
Our method is implemented on an NVIDIA GeForce RTX 4080S GPU. Initially, we employ the 3D supervoxel clustering method \cite{ADNET} to generate pseudo-masks as the supervision in the episode-based meta-learning task, and we follow the same pre-processing techniques as \cite{ADNET}. The experiments are conducted under the 1-way 1-shot condition. During inference, we randomly sample a scan from the source domain and select a middle slice containing the foreground as the support image, with the remaining slices as the query images. For all datasets, we train the model for 39K iterations, comprising 3000 iterations per epoch with the batch size set to 1. 
To comprehensively test the performance of our proposed model, we conduct bidirectional evaluations within each dataset. For instance, in the Cross-Modality dataset, we evaluate performance both on CT $\to$ MRI and MRI $\to$ CT directions.

Additionally, for the training of our model, the output size $N$ for the adaptive average pooling in FAM is set to $30^2$. In the multi-spectral decoupling of foreground features, we divide the frequency band into low, mid, and high frequencies with a ratio of 3:4:3.
We chose the Stochastic Gradient Descent (SGD) optimizer with an initial learning rate of 0.001, a momentum of 0.9 and a a decay factor of 0.95 every 1K iterations.

\subsection{Evaluation metric}
In order to evaluate the model under a uniform standard, we adopt the Sorensen-Dice coefficient \cite{ssl-alp}
that is commonly used in FSMIS tasks, as the evaluation metric. The Dice score is used to evaluate the overlap between the segmentation results and the ground truth, which can be denoted as 
\begin{equation}
    \text{DSC}(X, Y) = \frac{2 |X \cap Y|}{|X| + |Y|},
\end{equation}
where $X$ and $Y$ denote the two masks respectively, and the DSC denotes the Dice score ranges from 0 to 1, with 1 indicating complete overlap and 0 indicating no overlap.

\begin{table}[!t]
\centering
\fontsize{9}{11}\selectfont 
\begin{tabular}{>{\centering\arraybackslash}p{0.09\linewidth}>{\centering\arraybackslash}p{0.04\linewidth}>{\centering\arraybackslash}p{0.04\linewidth}>{\centering\arraybackslash}p{0.06\linewidth}|>{\centering\arraybackslash}p{0.06\linewidth}>{\centering\arraybackslash}p{0.06\linewidth}>{\centering\arraybackslash}p{0.06\linewidth}>{\centering\arraybackslash}p{0.06\linewidth}>{\centering\arraybackslash}p{0.06\linewidth}}
\toprule
\multirow{2}{*}{Baseline} & \multirow{2}{*}{CPG} & \multirow{2}{*}{FAM} & \multirow{2}{*}{MSF} & \multicolumn{5}{c}{CT $\to$ MRI}                \\
                          &                          &                      &                      & Liver & LK & RK & Spleen & Mean                 \\ \hline \rule{0pt}{2.5ex}
                          \checkmark &                          &                      &                      & 39.24 & 26.47 & 37.35 & 26.79 & \multicolumn{1}{l}{32.46} \\
                          \checkmark &                          \checkmark &                      &                      & 69.22& 49.52& 45.73& 51.41& \multicolumn{1}{l}{53.97} \\
                          \checkmark &                          \checkmark & \checkmark &                      & \underline{71.68}& \underline{55.45} & \underline{67.20} & \underline{53.75} & \multicolumn{1}{l}{\underline{62.02}} \\
                          \checkmark &                          \checkmark & \checkmark & \checkmark & \textbf{73.01} & \textbf{57.28} & \textbf{74.68} & \textbf{58.21} & \multicolumn{1}{l}{\textbf{65.79}} \\ \bottomrule
\end{tabular}
\caption{Ablation studies for the effect of each component in Dice score (\%).}
\label{ablation_main}
\end{table}

\subsection{Quantitative and Qualitative Results}
To demonstrate the effectiveness of our proposed method, we compare its performance with various FSMIS models, including PANet \cite{PANET}, SSL-ALPNet \cite{ssl-alp}, ADNet \cite{ADNET}, QNet \cite{qnet}, CATNet \cite{CATNET}, and RPT \cite{zhu2023few}. Additionally, two CD-FSS models, PATNet \cite{PATNET} and IFA \cite{IFA} are also used for comparison. Note that we evaluate CD-FSS models without fine-tuning.

As shown in Table \ref{ABD}, our proposed method significantly outperforms all existing FSMIS and CD-FSS models under both CT $\to$ MRI and MRI $\to$ CT directions. Specifically, in the CT $\to$ MRI direction, the Dice score reached 65.79\%, which is 2.78\% higher than the second-best method. More significantly, the proposed model exhibited an overall 7.46\% higher performance compared to the highest corresponding method in the MRI $\to$ CT direction. While PATNet performs 2.37\% better than our model in the liver category in the MRI $\to$ CT direction, it underperforms in smaller categories like RK, LK, and spleen. This discrepancy arises from the imbalanced foreground and background in medical images, which the CD-FSS models do not adequately address. In contrast, our model is better adapted to medical scenarios, resulting in a higher overall Dice score.

As depicted in Table \ref{CARDIAC}, Our method consistently performs exceptionally well on the Cross-Sequence dataset compared to other methods, achieving the highest Dice scores of 71.58\% and 61.39\% in both directions, surpassing the second-best method by 10.14\% and 1.33\%, respectively. In the LV-BP category under the LGE $\to$ b-SSFP scenario, our model even surpasses the second-best model by 21.29\%, reaching 86.64\%, which is comparable to FSMIS models’ segmentation accuracy in non-cross-sequence conditions.

For quantitative and qualitative results on the Cross-Institution dataset and visual segmentation results on three cross-domain datasets, please refer to the Supplementary Materials.
All experiments demonstrate that our FAMNet is a medical image segmentation model with excellent generalization capabilities and minimal data dependency.

\begin{table}[!t]
\centering
\resizebox{\linewidth}{!}{
\begin{tabular}{ccc|>{\centering\arraybackslash}p{0.08\linewidth}>{\centering\arraybackslash}p{0.08\linewidth}>{\centering\arraybackslash}p{0.08\linewidth}>{\centering\arraybackslash}p{0.08\linewidth}>{\centering\arraybackslash}p{0.08\linewidth}}
\toprule
\multicolumn{3}{c|}{\begin{tabular}[c]{@{}c@{}}Frequency band\end{tabular}} & \multicolumn{5}{c}{CT $\to$ MRI} \\ 
Low & Mid & High & Liver & LK & RK & Spleen & Mean \\ \hline
\rule{0pt}{2.5ex}
$\mathbin{-}$& $\mathbin{+}$& $\mathbin{-}$& \textbf{73.01}& \textbf{57.28}& \textbf{74.68}& 58.21& \textbf{65.79}\\
$\mathbin{-}$& $\mathbin{+}$& $\mathbin{+}$& 66.28& \underline{55.68}& 62.41& \textbf{60.79}& \underline{61.29}\\
$\mathbin{+}$& $\mathbin{+}$& $\mathbin{-}$& \underline{68.14}& 53.47& \underline{64.09}& 54.36& 60.02\\
 $\mathbin{+}$& $\mathbin{+}$& $\mathbin{+}$& 64.46& 48.77& 62.21& \underline{59.28}& 58.68\\

\bottomrule
\end{tabular}}
\caption{Ablation study (in Dice score \%) for the distinct attention weightings in DSFBs \& DAFBs. Given attention matrix $\mathbf{A}$, $\mathbin{+}$  indicates using $\mathbf{A}$ for attention weighting, and $\mathbin{-}$ indicates using 1-$\mathbf{A}$ for attention weighting. }
\label{attention weighting method}
\end{table}

\subsection{Ablation Studies}
\textbf{Effect of each component.}
In this section, we discuss the effect of each component. Table \ref{ablation_main} shows the contribution of each module to the overall model performance. Combined with CPG, the proposed FAM module significantly enhances baseline (PANet) performance by 29.56\%, primarily by mitigating overfitting to DSFBs and implementing inter-domain debiasing for the support and query features. Additionally, the MSF module aids in integrating the frequency-decoupled features from the FAM module, further suppressing DVI, and contributing an additional 3.77\% improvement in model performance.

\noindent \textbf{Distinct attention weightings in DAFBs \& DSFBs.}
In the FAM module, we apply distinct attention weighting methods to assign weights to features belonging to DSFBs and DAFBs. This approach reduces the model's dependency on the support-query correlation within DSFBs while enhancing its focus on DAFBs. Table \ref{attention weighting method} illustrates the ablation study for distinct attention weightings.
It is evident that using the uniform attention weighting method throughout leads to a significant drop in the Dice score, with a reduction of 7.11\%, and applying positive attention weighting to any DSFB results in a decrease in model performance. This decline is attributed to substantial overfitting caused by the attention mechanism to the source domain's support-query correlation, which weakens the model's ability to generalize to novel target domains.

For further discussion, we present extensive ablation experiments in the Supplementary Materials, which include more method comparisons and hyperparameter analysis.

\section{Conclusion}
In this paper, we have addressed a novel task: cross-domain few-shot medical image segmentation (CD-FSMIS). We proposed a Frequency-aware Matching Network (FAMNet), which comprises a Frequency-aware Matching (FAM) module to enhance the model’s generalization capabilities and reduce support-query bias by performing attention-based matching of the foreground features for specific frequency bands, which simultaneously handle intra-domain and inter-domain variations. Furthermore, we introduced a Multi-Spectral Fusion (MSF) module to integrate features decoupled by the FAM module and further suppress the detrimental impact of domain-variant information on the model’s robustness. Extensive experiments on three cross-domain datasets demonstrated the excellent generalization ability and data independence of the proposed method.

\section{Acknowledgment}
This work was partly supported by the National Natural Science Foundation of China (NSFC) under Grant Nos. 62371235, 62076132 and 62072246,  partly by the Key Research and Development Plan of Jiangsu Province (Industry Foresight and Key Core Technology Project) under Grant BE2023008-2.

\bibliography{aaai25}

\end{document}


\pdfobjcompresslevel=0
\pdfminorversion=5
\maketitle

\section{Ablation Studies}
Unless otherwise specified, all experiments are conducted using the same training setting and model configuration, consistent with the implementation details.

\begin{table}[h]
\fontsize{9}{11}\selectfont 
\begin{tabular}{>{\centering\arraybackslash}p{0.18\linewidth}|c|>{\centering\arraybackslash}p{0.07\linewidth}>{\centering\arraybackslash}p{0.07\linewidth}>{\centering\arraybackslash}p{0.07\linewidth}>{\centering\arraybackslash}p{0.07\linewidth}>{\centering\arraybackslash}p{0.07\linewidth}}
\toprule
Attention                       & Affected                       & \multicolumn{5}{c}{CT $\to$ MRI}      \\
Mechanisms & Fg Pixels & Liver & LK & RK & Spleen & Mean \\ \hline
\rule{0pt}{2.5ex}
\multirow{3}{*}{\begin{tabular}[c]{@{}c@{}}Hard \\ Attention\end{tabular}} & 20\%                           &       74.28&    56.49&    \underline{69.44}&        55.25&      63.87\\
                                & 50\%                           &       \underline{74.29}&    \textbf{58.47}&    66.45&        \underline{57.00}&      \underline{64.05}\\
                                & 80\%                           &       \textbf{76.44}&    56.06&    64.66&        53.54&      62.68\\ \hline 
\begin{tabular}[c]{@{}c@{}} \rule{0pt}{2.5ex}Soft \\ Attention\end{tabular}                  & 100\%                          &       73.01&    \underline{57.28}&    \textbf{74.68}&        \textbf{58.21}       &      \textbf{65.79}                           \\ \bottomrule
\end{tabular}
\caption{Comparison of hard attention at different dropout rates and soft attention in Dice score (\%).}
\label{softhardattention}
\end{table}

\subsection{Attention Mechanism in FAM}
In this section, we discuss the implementation of attention in the FAM module. In our experiments, the FAM module employs soft attention to weight the support and query foreground features, thereby mitigating the model's reliance on DSFBs. An alternative approach involves using hard attention to discard similar points directly. We conducted analytical experiments to analyze the feasibility of this method and compared the results with our soft attention approach. The comparative results in Table \ref{softhardattention} demonstrated the superiority of our method. Specifically, we selected the top {20\%, 50\%, 80\%} of points based on the computed similarity matrix and discarded them accordingly.

\subsection{Feasibility of Directly Discarding Frequency Bands}
In a previous section “Multi-Spectral Fusion (MSF) Module”, we concluded that directly discarding DSFBs is unreasonable. Table \ref{dropband} presents experimental evidence supporting this conclusion. The experiments were conducted using the FAM module but without the inclusion of the MSF module. We observed that discarding any one or more frequency bands adversely affects the model's performance compared to using all frequency bands, with a reduction of up to 3.90\%.

Based on these results, we conjecture that this outcome is related to the content-irrelevance of frequency decoupling. For instance, MRI images contain a large amount of fine textures that are absent in CT images, causing a domain shift. When discarding the high-frequency band, the fine texture information should be discarded as well. However, the high-frequency band also includes edge information, which plays a significant positive role in segmentation \cite{GMRD}, and this information remains consistent between CT and MRI (edge information of organs is similar in both modalities). Directly discarding the high-frequency band leads to the loss of both DVI and DII, resulting in the model learning in an information-deficient environment and thus decreasing segmentation performance. Furthermore, directly discarding a specific frequency band prevents the attention weights in that band from being trained, leading to over-fitting in the remaining frequency bands.

\begin{table}[!t]
\centering
\begin{tabular}{ccc|>{\centering\arraybackslash}p{0.08\linewidth}>{\centering\arraybackslash}p{0.08\linewidth}>{\centering\arraybackslash}p{0.08\linewidth}>{\centering\arraybackslash}p{0.08\linewidth}>{\centering\arraybackslash}p{0.08\linewidth}}
\toprule
\multicolumn{3}{c|}{\begin{tabular}[c]{@{}c@{}}Frequency band\end{tabular}} & \multicolumn{5}{c}{CT $\to$ MRI} \\ 
Low & Mid & High & Liver & LK & RK & Spleen & Mean \\ \hline
\rule{0pt}{2.5ex}
\checkmark & \checkmark & \checkmark & \underline{71.68}& \underline{55.45}& \textbf{67.20}& 53.75& \textbf{62.02}\\
 & \checkmark & \checkmark & 71.62& 54.89& 64.21& \underline{55.54}& \underline{61.57}\\
\checkmark & \checkmark &  & 70.77& \textbf{55.61}& \underline{66.64}& 53.12& 61.54\\ 
&  \checkmark &  & \textbf{71.71}& 50.05& 55.06& \textbf{55.66}& 58.12\\ 
\bottomrule
\end{tabular}
\caption{The impact of directly discarding a specific frequency band in Dice score (\%).}
\label{dropband}

\end{table}

\subsection{Performing Matching Exclusive to Specific Frequency Bands}
Instead of matching features across all frequency bands, this section focuses on matching exclusive to specific frequency bands. At least one DAFB is used to train the attention weights for support-query correlation to prevent overfitting issues, as detailed in the section “Distinct Attention Weightings in DAFBs \& DSFBs". The results of the study are shown in Table \ref{PARTIALLY_MATCHING}. When the low or high-frequency bands are not matched, the Dice score decreases by 2.59\% and 2.01\%, respectively, compared to matching across all frequency bands. Excluding both low and high-frequency bands from matching results in a minimum Dice score of 61.63\%. This decline occurs because the query sample may be out of distribution or the support prototype may not accurately represent the mean of a category due to potential intra-domain shifts between support and query samples, a phenomenon particularly evident in the 1-shot setting. These inter-domain variations have detrimental effects on the segmentation performance of the CD-FSMIS model. Figure \ref{fig:TSNE} illustrates the debiasing effect achieved by our proposed FAM module.

\begin{table}[t]
\centering
\begin{tabular}{ccc|>{\centering\arraybackslash}p{0.08\linewidth}>{\centering\arraybackslash}p{0.08\linewidth}>{\centering\arraybackslash}p{0.08\linewidth}>{\centering\arraybackslash}p{0.08\linewidth}>{\centering\arraybackslash}p{0.08\linewidth}}
\toprule
\multicolumn{3}{c|}{\begin{tabular}[c]{@{}c@{}}Frequency band\end{tabular}} & \multicolumn{5}{c}{CT $\to$ MRI} \\ 
Low & Mid & High & Liver & LK & RK & Spleen & Mean \\ \hline
\rule{0pt}{2.5ex}
\checkmark & \checkmark & \checkmark & \underline{73.01}& \textbf{57.28}& \textbf{74.68}& \textbf{58.21}& \textbf{65.79}\\
 & \checkmark & \checkmark & 72.62& 55.68& \underline{67.25}& 57.26& 63.20\\
\checkmark & \checkmark &  & \textbf{74.01}& \underline{57.08}& 66.62& \underline{57.40}& \underline{63.78}\\ 
&  \checkmark &  &  72.90&  56.00&  64.26&  53.05&  61.63\\ 
\bottomrule
\end{tabular}
\caption{Ablation study (in Dice score \%) for the impact of matching is exclusive to specific frequency bands. $\checkmark$ signifies feature matching in the corresponding frequency band.}
\label{PARTIALLY_MATCHING}
\end{table}

\subsection{Impact of Different Frequency Band Division Ratios}
In this section, we discuss the impact of different frequency band division ratios on model performance. The division ratio determines our segmentation of DSFBs and DAFBs. In the FAM module, we apply different attention weightings to the image features belonging to DSFBs and DAFBs based on this segmentation. In the MSF module, we extract DII from features in DAFBs to extract residual DII from DSFBs, while simultaneously suppressing DVI in the final module output. Table \ref{DIVISIONRATIO} shows the impact of different division ratios.
From Table \ref{DIVISIONRATIO}, we observe that our model's performance peaks with a Dice score of 66.29\% when the low:mid:high ratio is 3.5:3:3.5. When the proportion of the mid-frequency band exceeds 30\%, the model's performance shows a decreasing trend as the mid-frequency band increases. We attribute this to the blurred boundary between DSFBs and DAFBs. During frequency band division, DAFBs inevitably include some frequency domain information from DSFBs. As the bandwidth of the mid-frequency band increases, DAFBs contain more domain-specific frequency signals. This deviates from the design intent of the FAM and MSF modules. In the FAM module, certain support-query matching relationships from DSFBs are also reinforced, which could be crucial for optimization during source domain training, leading to overfitting of source domain information. In the MSF module, since the mid-frequency band contains DVI, its guiding significance in the cross-attention extraction process decreases, resulting in certain DVI in DSFBs being enhanced rather than suppressed. When the proportion of the mid-frequency band falls below 30\%, the model's performance declines sharply due to insufficient information in the mid-frequency band, rendering both the FAM and MSF modules inadequately trained.

\begin{table}[!t]
\centering
\resizebox{\columnwidth}{!}{
\begin{tabular}{ccc|>{\centering\arraybackslash}p{0.08\linewidth}>{\centering\arraybackslash}p{0.08\linewidth}>{\centering\arraybackslash}p{0.08\linewidth}>{\centering\arraybackslash}p{0.08\linewidth}>{\centering\arraybackslash}p{0.08\linewidth}}
\toprule
\multicolumn{3}{c|}{\begin{tabular}[c]{@{}c@{}}Division Ratio \\(Low:Mid:High)\end{tabular}} & \multicolumn{5}{c}{CT $\to$ MRI} \\ 
Low & Mid & High & Liver & LK & RK & Spleen & Mean \\ \hline
\rule{0pt}{2.5ex}
2.0 & 6.0 & 2.0& \textbf{75.11}& 56.49 & 66.35 & 57.27  & 63.81 \\
2.5 & 5.0 & 2.5&       73.19&       57.06&       70.75&        56.06&       64.27\\
3.0 & 4.0 & 3.0& 73.01& 57.28 & \textbf{74.68}& \underline{58.21}& 65.79 \\
3.5 & 3.0 & 3.5&       72.75&       \textbf{60.52}&       \underline{74.44}&        57.44&       \textbf{66.29}\\
4.0 & 2.0 & 4.0&       \underline{74.01}&       \underline{59.35}&       72.77&        57.37&       \underline{65.88}\\ 
4.5& 1.0& 4.5&  68.68&  55.77&  68.42&   \textbf{60.31}&  63.30\\ \bottomrule
\end{tabular}}
\caption{Ablation study (in Dice score \%) for the impact of different frequency band division ratios.}
\label{DIVISIONRATIO}
\end{table}

\subsection{Impact of Foreground Pixel Number $N$}
\begin{figure}[H]
    \centering
    \includegraphics[width=0.95\linewidth]{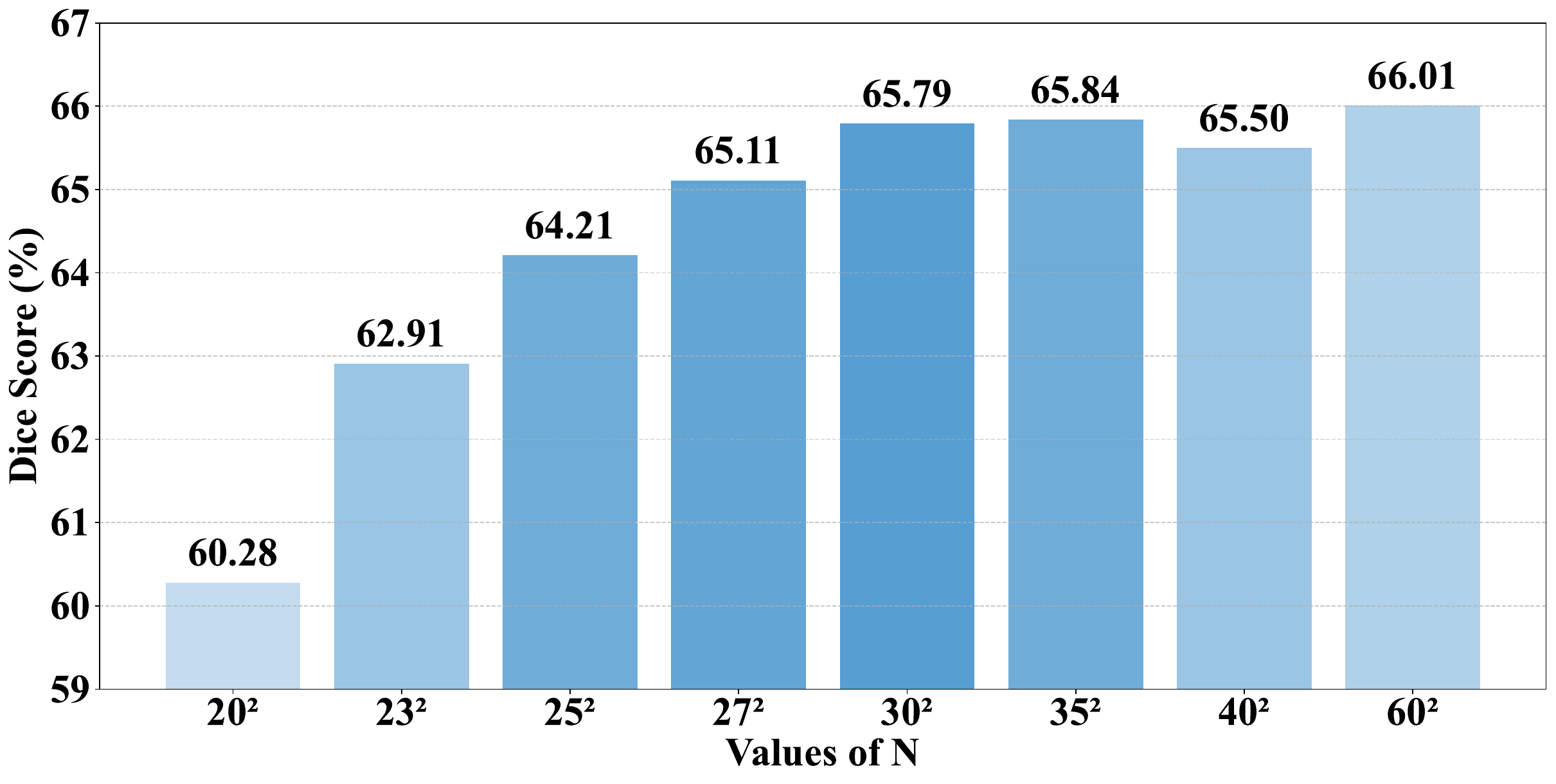}
    \caption{Ablation study (in Dice score \%) for the impact of different $N$ values.}
    \label{fig:hyperN}
\end{figure}

In this section, we analyze the fixed number $N$ of foreground pixels, which standardizes the number of foreground pixels through the adaptive average pooling. In our experiment, we selected $N$ as $30^2$. We further vary $N$ and evaluate the model's performance in CT $\to$ MRI direction, as illustrated in Figure \ref{fig:hyperN}. It can be observed that as $N$ increases, the Dice score initially rises steadily and then fluctuates around a certain value. When $N$ is $60^2$, the model achieves the highest Dice score, but the difference compared to the reported results is minimal. Meanwhile, since the value of $N$ is directly related to the subsequent network design, a large $N$ introduces a significant number of unnecessary parameters and computational load. Hence, it is crucial to choose a moderate value of $N$ to maintain our model's efficiency.

\begin{table*}[!t]
\centering
\begin{tabular}{c|c|cccc|cccc}
\toprule 
\rule{0pt}{2.5ex}
\multirow{2}{*}{Method} & \multirow{2}{*}{Ref.} & \multicolumn{4}{c|}{\textbf{Prostate UCLH $\to$ NCI}} & \multicolumn{4}{c}{\textbf{Prostate NCI $\to$ UCLH}} \\ 
\cline{3-10} 
\rule{0pt}{2.5ex}
                        &                       & \multicolumn{1}{c|}{Bladder} & \multicolumn{1}{c|}{CG} & \multicolumn{1}{c|}{Rectum} & Mean  & \multicolumn{1}{c|}{Bladder} & \multicolumn{1}{c|}{CG} & \multicolumn{1}{c|}{Rectum} & Mean  \\ \hline
\rule{0pt}{2.5ex}
 PANet                  & ICCV'19               & 51.85& 38.89& 36.28& 42.34& 49.90& 36.27& 43.58& 43.25\\
 SSL-ALP                & TMI'22                & \textbf{54.86}& 42.59& 41.65& 46.37& 66.6& 43.12& 56.94& 55.55\\
 ADNet                  & MIA'22                & \underline{53.92}& 46.11& 42.26& 47.43& 62.84& 51.34& 60.45& 58.21\\
 QNet                   & IntelliSys'23         & 39.66& 35.71& 34.25& 36.54& 41.22& 43.21& 39.19& 41.21\\
 CATNet                 & MICCAI'23             & 47.61& 45.29& 42.51& 45.14& 51.98& 48.81& 55.01& 51.93\\
 RPT                    & MICCAI'23             & 52.17& 41.22& \underline{46.93}& 46.77& 62.35& \underline{51.66}& \underline{64.63}& 59.55\\
 PATNet                 & ECCV'22               & 50.04& \textbf{52.71}& 44.57& \underline{49.10}& \underline{69.91}& \textbf{53.02}& 62.79& \underline{61.91}\\ 
 IFA                    & CVPR'24               & 41.02& 39.15& 36.45& 38.87& 52.54& 39.85& 34.27& 42.22\\ \hline
 \rule{0pt}{2.5ex}
  \textbf{Ours}        & \textemdash            &                            53.06&                             \underline{48.48}&                         \textbf{52.57}&       \textbf{51.37}&\textbf{77.64}&48.73& \textbf{68.91}&\textbf{65.09}\\ 
\bottomrule 
\end{tabular}
\caption{Quantitative comparison of different methods Dice score (\%) on the Cross-Institution Dataset. The best value is shown in bold font, and the second best is underlined.}
\label{PROSTATE}
\end{table*}

\section{Quantative and Qualitative Result on the Cross-Institution Dataset}
As shown in Table \ref{PROSTATE}, our FAMNet also excels on the Cross-Institution dataset, achieving a 2.27\% and 3.18\% higher Dice score in the UCLH $\to$ NCI and NCI $\to$ UCLH directions, respectively, compared to the second-best method. The performance improvement is attributed to FAMNet's ability to better address the variations arising from the use of instruments of different models or different manufacturers across institutions. These discrepancies typically result in differences in image contrast, signal-to-noise ratio, and other factors, which are particularly pronounced in the frequency domain, especially in the high and low-frequency bands \cite{contrast_in_freq,SNR}. FAMNet effectively mitigates the impact of these domain shifts on segmentation performance, thereby enhancing its robustness and accuracy across different institutional data.

\begin{figure}[!t]
    \centering
    \includegraphics[width=\linewidth]{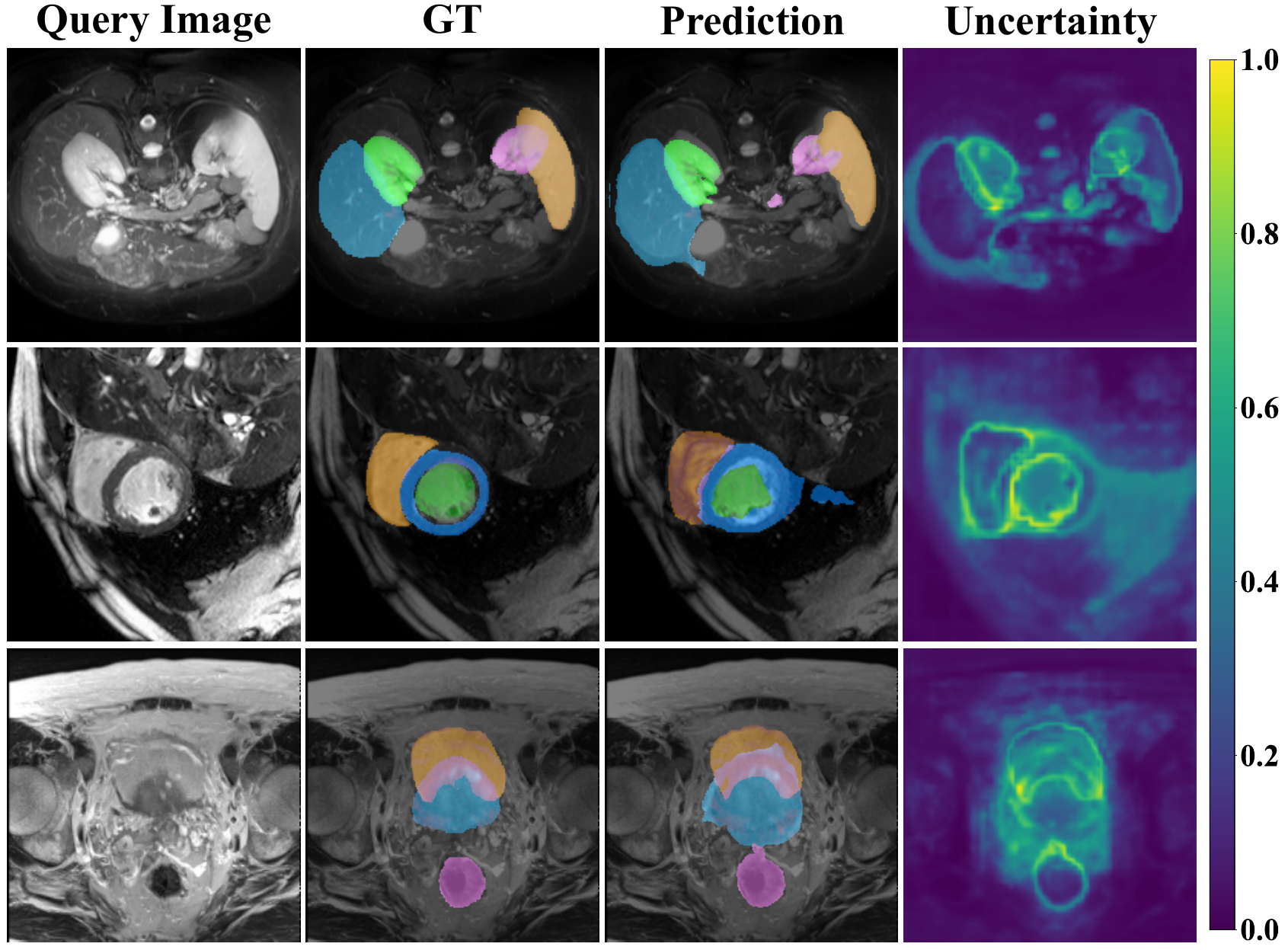}
    \caption{Illustration of query images, ground truths, predictions, and uncertainty maps in the directions of CT $\to$ MRI, LGE $\to$ b-SSFP, NCI $\to$ UCLH by FAMNet.}
    \label{fig:uncertainty}
\end{figure}

\begin{figure}[!t]
    \centering
    \includegraphics[width=\linewidth]{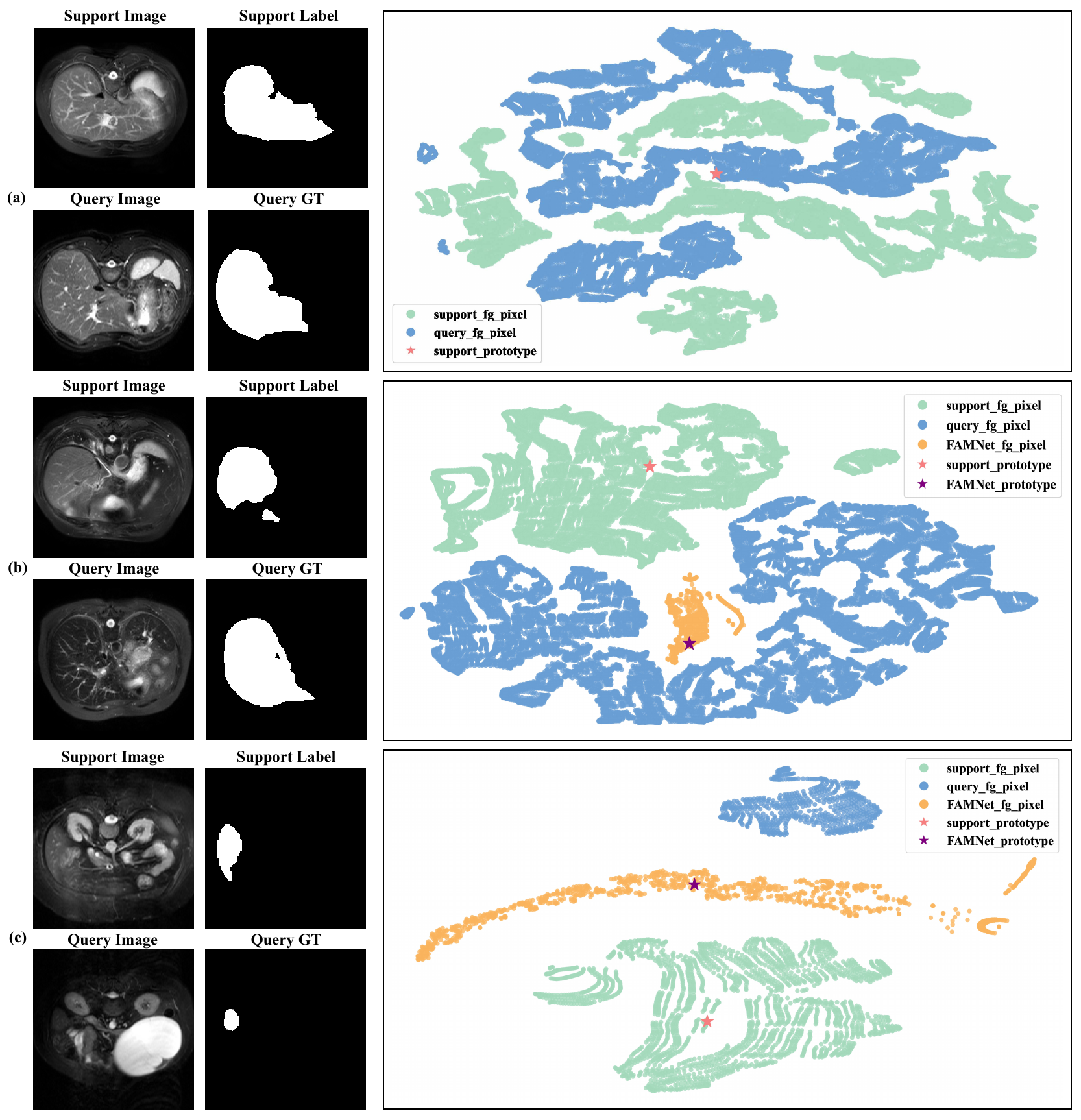}
    \caption{T-SNE visualization for intra-domain variations, \textit{i.e.}, support-query bias, and debiasing capability of FAMNet. 'fg' represents the foreground. In (b), the foregrounds in the support and query images exhibit significant differences in brightness and contrast, with notable discrepancies in texture and structural details as well. In (c), there is a pronounced size imbalance in the foregrounds, accompanied by substantial variations in brightness.}
    \label{fig:TSNE}
\end{figure}

\section{Visualization}
\subsection{Uncertainty Maps}
Figure \ref{fig:uncertainty} illustrates three segmentation examples by FAMNet from the CD-FSMIS task in the directions of CT $\to$ MRI, LGE $\to$ b-SSFP, and NCI $\to$ UCLH. By observing the predictions and uncertainty maps, it is evident that the edges of the target regions are typically highlighted. This indicates a higher model uncertainty in segmenting these areas, due to the differences in edge depiction across domains, \textit{e.g.}, CT images generally have sharper edges compared to MRI images. Additionally, the model exhibits greater uncertainty when segmenting smaller targets compared to larger organs. This is attributed to the varying focus on detailed regions by different imaging techniques. Furthermore, when multiple targets are in close proximity or intersect, there is an increased uncertainty, which may lead to false segmentation.

\subsection{FAMNet's Capability to Mitigate Intra-domain Variations}
Figure \ref{fig:TSNE} visualizes the capability of our FAMNet to significantly narrow the support-query bias using t-SNE \cite{TSNE}. As shown in Figure \ref{fig:TSNE}(a), typical support and query samples are closely situated and highly intermixed in the feature space, and the support prototype effectively represents the mean of the query foreground's overall distribution, resulting in excellent segmentation performance. Figure \ref{fig:TSNE}(b)(c) illustrates that when there is a moderate or significant bias between support and query samples, the support prototype deviates from the cluster of query foreground pixels in the feature space. This deviation reduces the representational capability of the support prototype, thereby degrading the performance of the cosine similarity-based mask calculation method. FAMNet calculates a new prototype through support-query matching, correcting the bias of the support prototype, thereby reducing or eliminating the detrimental impact of intra-domain variation on segmentation results.

\bibliography{aaai25}